\documentclass[12pt]{ut-thesis}
\usepackage{tabu}
\usepackage{url}
\usepackage{latexsym}
\usepackage{comment}
\usepackage{tabu}
\usepackage{amsmath}
\usepackage{tablefootnote}
\usepackage{multirow}
\usepackage{algorithm}
\usepackage{algorithmic}
\usepackage{bbm}
\usepackage{enumitem}
\usepackage{graphicx}
\usepackage{setspace}
\graphicspath{{./figures/}}
\usepackage[round]{natbib}
\newcommand{\rowgroup}[1]{\hspace{-1em}#1}
\usepackage{booktabs} 
\usepackage{pbox}
\usepackage{array}
\usepackage{hanging}
\usepackage{float}

\degree{Master of Science}
\department{Computer Science}
\gradyear{2016}
\author{Krish Perumal}
\title{Semi-supervised and unsupervised methods for categorizing posts
in Web discussion forums}

\begin{document}

\begin{preliminary}

\maketitle


\begin{abstract}
Web discussion forums are used by millions of people worldwide to
share information belonging to a variety of domains such as automotive
vehicles, pets, sports, etc. They typically contain posts that fall
into different categories such as \textit{problem}, \textit{solution},
\textit{feedback}, \textit{spam}, etc. Automatic identification of these categories can aid information
retrieval that is tailored for specific user requirements. Previously, a number of supervised methods have attempted to
solve this problem; however, these depend on the availability of abundant
training data. A few existing unsupervised and semi-supervised
approaches are either focused on identifying a single category or do not report
category-specific performance. In contrast, this work proposes unsupervised and
semi-supervised methods that require no or minimal training data to achieve this objective without
compromising on performance. A fine-grained analysis is also carried
out to discuss their limitations. The proposed methods are based on
sequence models (specifically, Hidden Markov Models) that can model
language for each category using word and part-of-speech
probability distributions, and manually specified features. Empirical evaluations
across domains demonstrate that the proposed methods are better suited
for this task than existing ones.
\end{abstract}





\begin{acknowledgements}
I thank my advisor, Prof. Graeme Hirst, whose valuable
guidance, feedback and attention to detail was pivotal for the completion of this work. I also thank my
second reader, Prof. Gerald Penn, for assessing my work and
providing me valuable comments.

I am grateful to Dr. Afsaneh Fazly who first introduced me to this
work, and advised me on possible research directions to explore at
every step of this work. I also thank Brandon Seibel and Alex Minnaar
for assisting me during my visits to Verticalscope Inc. I sincerely
thank Verticalscope Inc. for allowing me to access
their data for this research.

I sincerely thank the Natural Sciences and Engineering Research Council of
Canada for supporting this research through the NSERC-ENGAGE grant
(no.\@ EGP 477227-14). I am also grateful to Mohamed Abdalla for
assisting me with some of the implementations mentioned in this work.

I thank my parents who worked tirelessly to provide me a great
education. I also thank my friends and well-wishers at the Department
of Computer Science who assisted me (both academically and otherwise) during my
time here -- Kaustav Kundu, Dave Fernig, Ivan Vendrov, Nona Naderi,
Jamie Kiros, Aida Nematzadeh, Aditya Bhargava, Sean Robertson, Patricia
Thaine, Katie Fraser and Tong Wang. I also specially thank my best friends back
in India --- Sidhant Sharma, Kartik Arunachalam, Nishant
Gupta, and Vishi Jhalani.

\end{acknowledgements}

\tableofcontents




\end{preliminary}



\chapter{Introduction}
The Internet contains a wide range of user-generated content in the
form of blogs, discussion forums, social media posts, digital media,
etc. These enable users to exchange information in a manner less
formal and more personalized than centralized information sources such as
government agencies, media houses, and educational and research institutes. Among
these, Web discussion forums are platforms where people
converse with one another to collaboratively solve problems and
discuss issues. These forums might encompass a wide range of topics
(e.g., Yahoo Answers\footnote{\url{http://answers.yahoo.com}}) or be
limited to a narrow domain (e.g., JeepForum\footnote{\url{http://www.jeepforum.com/}}). The former kind of
forums are typically organized into topic hierarchies, essentially
reducing them to forums of the latter kind. For example, Yahoo Answers
consists of topics such as \textit{Arts and Humanities}, \textit{Health},
\textit{Family and Relationships}, etc. Further, \textit{Family and
Relationships} contains sub-topics such as \textit{Family}, \textit{Friends},
\textit{Marriage and Divorce}, etc. Such a hierarchy enables easier
navigation for users who wish to seek or provide information about a
specific topic of their interest. Within each topic, forums consist of
individual conversations, called threads, each containing
multiple user messages, called posts.

\begin{table}[t!]
\scriptsize
\centering
\resizebox{\columnwidth}{!}{
{\tabulinesep=2mm
\begin{tabu}{>{\quad} l p{2cm}}
\hline
\textbf{Post} & \textbf{Purpose} \\
\hline
\rowgroup{User \textit{15JKU}:} \\
\parbox{11cm}{Hey Guys,
\\
Im fairly new to the jeep world. Im looking to get 35s with either 18s or 20s as it will be more of a daily driver and sometimes go mudding. I live in miami so I'm not really concerned on any dinks and dangs on my wheels. 
\\
I have a buddy that can get me set up with brand new Nitto Trail's for an awesome price. My only concern is how they will perform in mud? Also, how loud would they for a daily driven jeep?
\\
Also, would A/T tires work for mudding? No I assume. What tires are worth getting without breaking bank?
\\
Thanks in advance!}
& \vspace{-2cm}\parbox{2cm}{Posting\\a problem} \\
\hline
\rowgroup{User \textit{mschi772}:} \\
\parbox{11cm}{
You want 35's on 18-20" wheels that are good in the mud, good daily drivers that aren't too loud, and won't break the bank? Why not ask for good snow and ice performance, too? You need to more accurately convey what your true priorities are here because you're asking for too much from one tire.
}
& \vspace{-1.2cm}\parbox{2cm}{Requesting\\clarification} \\
\hline
\rowgroup{User \textit{15JKU}:} \\
\parbox{11cm}{
Just asking if anyone knows how loud they are. My main concern is how they'll do on mud and if i should go with different tire
}
& \vspace{-0.8cm}\parbox{2cm}{Clarifying to\\previous user} \\
\hline
\rowgroup{User \textit{mschi772}:} \\
\parbox{11cm}{
Nitto Trail Grapplers are a "classic" MT design. You'll see nearly identical MT tread patterns from many other companies (BFG MT, Firestone MT, Toyo MT, Cooper STT, etc). This is a very popular design for people who frequently go offroading but want to maintain some street manners. They do fine in mud. There are better tires for mud, but they would be loud and handle poorly on the street as well has get worn-out VERY quickly on the street. If you've got access to a great deal on them, go for it.
}
& \vspace{-1.6cm}\parbox{2cm}{Providing\\a solution} \\
\hline
\rowgroup{User \textit{JcArnold}:} \\
\parbox{11cm}{
I've got 37" trails and they are not noisy. I live in Colorado so I don't know about mud but they are great tires in the rocks and snow.
}
& \vspace{-0.8cm}\parbox{2cm}{Providing\\a solution} \\
\hline
\rowgroup{User \textit{Pedro7}:} \\
\parbox{11cm}{
If you are concerned about noise, don't get a mud tire. If you're concerned about mud performance, get a mud tire. Every mud tire is going to be somewhat noisy, especially when they wear. Take the good with the bad.
\\
I have nittos. They sound like a mud tire on the road, but I've had worse....and yes, they work well in mud. AT tires don't work well in mud
\\
If you want 18-20s, you will break the bank. Nittos are a very top of the line MT. If your buddy can get you a deal, get them.
\\
Cheaper tire will be terrible as they wear, worse on the road, rain, etc and will be louder.
\\
See what I'm getting at? There is no perfect tire for every situation, but, nittos are close.
}
& \vspace{-2.2cm}\parbox{2cm}{Providing\\a solution} \\
\hline
\rowgroup{User \textit{15JKU}:} \\
\parbox{11cm}{
Thanks guys! Truly appreciate it. I'll go with the Nitto grapplers.
}
& \vspace{-0.7cm}\parbox{2cm}{Providing\\feedback} \\
\hline
\end{tabu}
}
}
\caption{\label{table:example-thread} Example discussion
  forum (source:
  \protect\url{http://www.jeepforum.com/forum/f15/tire-recommendations-3455674/})
  with the manually identified purpose of each post.}
\end{table}
\clearpage

An example discussion forum thread is shown in Table
\ref{table:example-thread}. Here, user \textit{15JKU} (called
\textit{original poster} from here onward) initially asks for advice on whether 35's model tires
on 18-20-inch wheels are good for daily driving as well as mudding (a hobby of driving jeeps on muddy off-road
surfaces). The user also does not want the wheels to make much
noise. User \textit{mschi772} responds that the \textit{original poster} is expecting
too much from a single tire, and requests clarification on the
user's priorities. The original poster clarifies that he/she wants to know
how loud the models of wheels and tires are, and that the priority is
suitability for mudding. User \textit{mschi772} proposes another model called
Nitto Trail Grapplers which are better for mudding, but would make noise and wear out quickly on
streets. User \textit{JcArnold} responds that he/she has 37-inch trails which
work well in rock and snow. User \textit{Pedro7} joins the
conversation by asking the original poster to not go for mud tires if
noise is a concern. The user recommends Nittos as the best
possible solution, but warns against expecting a perfect tire for every
situation. Finally, the \textit{original poster} provides feedback by thanking
everyone and announcing that he/she is choosing Nittos.

Table \ref{table:example-thread} also contains a column (which is not
part of the original forum) mentioning the manually identified purpose of each post in the thread. With this information, a user
seeking a solution to a similar problem need only read three out of
six posts replying to the first post. Without such information, the user must read the entire
thread. This problem becomes much more pronounced in cases where
threads contain tens or hundreds of posts, and reading the entire
thread becomes impractical (unless one participates in the thread
conversation from the beginning). For example, \url{http://www.jeepforum.com/forum/f15/mud-tires-119948/} contains
more than 500 posts discussing popular brands of tires. Most of these
posts involve off-topic personalized discussions. In such
cases, the purpose of each post can guide the user towards useful
posts (i.e., containing solutions) and away from trivial posts (i.e.,
containing feedback or off-topic discussions). Moreover, current
information retrieval techniques return entire threads as results to search queries. But by being sensitized to these
annotations, they can return targeted results containing only relevant posts
instead of entire threads. Further, user-contributed information
contained in these forums can be better structured and contribute
towards the development of domain-specific knowledge bases. With these
motivations in mind, this work aims to automatically annotate each post in a discussion forum with its purpose in the conversation thread.

The problem described is neither a novel nor a neglected one in the field of
computational linguistics (as will be demonstrated in the discussion of related work in the following chapter). 
It is closely related to the problem of \textit{dialogue act tagging},
which is defined as the identification of the meaning of an
utterance at the level of illocutionary force \citep{Stolcke2000},
i.e., an utterance could be identified as falling into one or more categories
such as \textit{problem}, \textit{solution}, \textit{clarification},
\textit{feedback}, \textit{command}, \textit{request}, etc. Most of
the previous work has concentrated on supervised machine
learning methods which make use of manually annotated data in order to predict the
annotations of unseen data. In contrast, this paper discusses novel approaches
using minimal (semi-supervised) or no manually annotated data
(unsupervised). Some previous work on semi-supervised and
unsupervised methods exists; however, this research paper will
empirically demonstrate (in section 5) that the
proposed methods perform better.

The main contributions of this work are the following.
\begin{itemize}[noitemsep]
\item Summarizing existing work on categorizing discussion
  forum posts and discussing their limitations.
\item Proposing novel methods based on sequence models for categorizing discussion forum posts with minimal or no annotated data.
\item Developing an annotated dataset of discussion forums from a hereto neglected
  automotive domain.
\item Conducting experiments to analyze the performance of existing
  and proposed methods on datasets belonging to different domains.
\end{itemize}


\chapter{Related Work}
The problem of identifying the purpose or intention of each post in a discussion forum thread has been extensively tackled in previous literature. However, there is no unanimously agreed-upon set of tags to identify, because they depend on the final objective of the tagging process. For example, the objective of an answer retrieval system is better achieved by concentrating on identifying \textit{Question} and \textit{Answer} posts alone, whereas the objective of an answer quality assessment system is fulfilled by additionally identifying \textit{Positive/Negative Feedback} posts. Most of the previous work has concentrated on tackling these kind of dialogue categories, and uses the term \textit{dialogue act tagging}. Also, some previous work has named categories specific to the target domain. For example, a forum on the medical domain may typically consist of posts explaining medical conditions and those providing treatment options, hence identifying categories such as \textit{Medical Problem} and \textit{Treatment}, whereas a forum on the computer-related technical domain may consist of categories such as \textit{Problem:\@ Hardware}, \textit{Problem:\@ Software}, \textit{Solution:\@ Install} and \textit{Solution:\@ Search}. This research paper does not restrict itself solely to \textit{dialogue act tagging}; neither does it address the classification of categories for only a specific domain. Hence, it uses the general term \textit{forum post categorization}.

\begin{table}[t!]
\small
\centering
\resizebox{\columnwidth}{!}{
{\tabulinesep=1mm
\begin{tabu}{p{15cm}}
\hline
\textbf{\cite{Ding2008}}\\
\hspace{1em}Tagset: \textit{Question Context}, \textit{Answer}\\
\hspace{1em}Classifier: CRF\\
\hspace{1em}Features: Similarity, structural, discourse, lexical\\
\hspace{1em}Dataset: TripAdvisor (travel domain)\\
\hline
\textbf{\cite{Sondhi2010}}\\
\hspace{1em}Tagset: \textit{Medical Problem}, \textit{Treatment}\\
\hspace{1em}Classifiers: CRF, SVM\\
\hspace{1em}Features: Semantic, structural\\
\hspace{1em}Dataset: HealthBoards (medical domain)\\
\hline
\textbf{\cite{Wang2010a}}\\
\hspace{1em}Tagset: \textit{Problem -- Hardware, Software, Media, OS, Network, Programming}; \\
\hangpara{2em}{0}\textit{Solution -- Documentation, Install, Search, Support};\\
\hangpara{2em}{0}\textit{Miscellaneous -- Spam, Other}\\
\hspace{1em}Classifiers: SVM, naive Bayes\\
\hspace{1em}Features: Bag-of-words\\
\hspace{1em}Dataset: CNET (computer-related technical domain)\\
\hline
\textbf{\cite{Kim2010}}\\
\hspace{1em}Tagset: \textit{Question}, \textit{Question-Add}, \textit{Question-Confirmation}, \textit{Question-Correction},\\
\hangpara{2em}{0}\textit{Answer}, \textit{Answer-Add}, \textit{Answer-Confirmation}, \textit{Answer-Correction}, \textit{Answer-Objection}, \textit{Resolution}, \textit{Reproduction}, \textit{Other}\\
\hspace{1em}Classifier: CRF\\
\hspace{1em}Features: Lexical, structural, post context, semantic\\
\hspace{1em}Dataset: CNET (computer-related technical domain)\\
\hline
\textbf{\cite{Qu2011}}\\
\hspace{1em}Tagset: \textit{Problem}, \textit{Solution}, \textit{Good Feedback}, \textit{Bad Feedback}\\
\hspace{1em}Classifier: HMM\\
\hspace{1em}Features: Bag-of-words\\
\hspace{1em}Dataset: Oracle database (computer-related technical domain)\\
\hline
\textbf{\cite{Catherine2012}}\\
\hspace{1em}Tagset: \textit{Answer}\\
\hspace{1em}Classifier: SVM\\
\hspace{1em}Features: Structural, syntactic, author authority, post ratings\\
\hspace{1em}Dataset: Apple (computer-related technical domain)\\
\hline
\textbf{\cite{Bhatia2012}}\\
\hspace{1em}Tagset: \textit{Question}, \textit{Repeat Question}, \textit{Clarification}, \textit{Solution}, \textit{Further Details}, \\
\hangpara{2em}{0}\textit{Positive Feedback}, \textit{Negative Feedback}, \textit{Spam}\\
\hspace{1em}Classifiers: SVM, logit model\\
\hspace{1em}Features: Structural, content, sentiment, number of posts by user, user authority\\
\hspace{1em}Datasets: Ubuntu (computer-related technical domain),\\
\hangpara{2em}{0}TripAdvisor-NYC (travel domain)\\
\hline
\end{tabu}
}
}
\caption{\label{table:existing-supervised} Existing supervised methods for categorizing forum posts.}
\end{table}

\clearpage

\section{Supervised Methods}
Supervised machine learning methods use previously labeled data for training, in order to predict the categories assigned to unseen data. Related previous work on classification of categories of discussion forum posts has largely focused on the application of these methods. In particular, most of the work has concentrated on the computer-related technical domain. \cite{Catherine2012} employed Support Vector Machines (SVMs) to extract \textit{Answer} posts in a thread (assuming that the first post in the thread is a \textit{Question}). They used a number of structural and syntactic features, in addition to forum-specific features such as author authority\footnote{According to \cite{Catherine2012}, author authority is a numerical or categorical value that is indicative of an author's level of expertise in the context of the forum.} and post ratings. Their methods were evaluated on a corpus of Apple discussion forums\footnote{\url{https://discussions.apple.com}}. \cite{Bhatia2012} used supervised machine learning algorithms (i.e., SVMs, logit model classifier, naive Bayes, etc.\@) to classify forum posts into eight categories --- \textit{Question}, \textit{Repeat Question}, \textit{Clarification}, \textit{Further Details}, \textit{Solution}, \textit{Positive Feedback}, \textit{Negative Feedback}, and \textit{Junk}. They evaluated their methods on a dataset of the Ubuntu forums\footnote{\url{http://ubuntuforums.org}}. \cite{Qu2011} used Hidden Markov Models (HMMs) to classify forum posts into four categories --- \textit{Problem}, \textit{Solution}, \textit{Good Feedback} and \textit{Bad Feedback}. They evaluated their methods on the Oracle database support forums\footnote{\url{https://community.oracle.com/community/database/}}. Similarly, \cite{Wang2010a} attempted to identify \textit{Problem} and \textit{Solution} posts in the CNET forums\footnote{\url{http://forums.cnet.com}} dataset, but with more fine-grained categories based on the types of \textit{Problem} posts (i.e., \textit{Hardware}, \textit{Software}, \textit{Media}, \textit{OS}, \textit{Network}, and \textit{Programming}) and \textit{Solution} posts (i.e., \textit{Documentation}, \textit{Install}, \textit{Search}, and \textit{Support}). \cite{Kim2010} worked on the same dataset, and attempted to classify posts into 12 categories that are similar to the ones used by \cite{Bhatia2012}. Additionally, they tagged the links between posts, i.e., identifying which post is a reply to which other post. For both tasks, they reported the best performance using Conditional Random Fields (CRFs). \cite{Wang2011} went one step further by jointly classifying both posts and the links between them. They used two different methods: (1) composition of results from both tasks done separately, and (2) combination of post and link tag sets in a single task. Two other papers reported work on forums on the travel and medical domains. \cite{Ding2008} used CRFs to identify \textit{Answer} posts and the context in which they answered the \textit{Question} post. However, they did not attempt to identify \textit{Question} posts, because they were assumed to be known beforehand. Their techniques were evaluated on a corpus of the TripAdvisor forums\footnote{\url{http://www.tripadvisor.com/ForumHome}}. \cite{Sondhi2010} used CRFs and SVMs with various semantic and structural features to identify \textit{Medical Problem} and \textit{Treatment} in the HealthBoards forums\footnote{\url{http://www.healthboards.com}}. A summary of all these methods is presented in Table \ref{table:existing-supervised}. A major drawback of these approaches is that they are constrained by the requirement of manually annotated data for training, and are limited in applicability to the domains they are trained on.

\section{Unsupervised Methods}
Unsupervised methods identify unlabeled clusters of data, each of which could potentially be mapped to a target category that one wants to identify. These methods use a task-dependent measure of similarity to identify whether two input units should belong to the same cluster or not, and in some cases, also model the interactions between the clusters. In contrast to supervised techniques, they require no labeled data; hence, they are not limited in applicability to a specific domain. To the best of our knowledge, three unsupervised techniques have been previously proposed for categorization of posts in Web forums. \cite{Cong2008} used labeled sequential patterns to identify \textit{Question} posts, followed by a graph-based propagation method to extract corresponding \textit{Answer} posts. The question detection phase was supervised, whereas answer extraction was unsupervised. The graph-based propagation used language models and author authority in order to assign scores to the links (edges) between posts (nodes). The method was evaluated on a corpus of forum threads on the travel domain. \cite{Deepak2014} identified \textit{Solution} posts using a translation-based model that leverages lexical correlations between \textit{Problem} and \textit{Solution} posts. \cite{Joty2011} used a combination of HMMs and Gaussian Mixture Models (GMMs) in order to classify forum posts into 12 dialogue act categories. In addition to word n-grams, they used some structural features such as the chronological position of a post in the thread, the number of tokens in the post, and author identity. Both these papers reported results on corpora of forums on the computer-related technical domain (i.e., Apple discussion forums and Ubuntu forums). Other unsupervised techniques have been employed for the related tasks of \textit{dialogue act classification} in spoken dialogue systems \citep{Crook2009} and Twitter conversations \citep{Ritter2010}. Although they worked specifically on genres of text that are very different from Web forums, they can potentially inspire future approaches tailored for Web forums. All these unsupervised approaches ignored the evaluation of category-wise classification. Instead, they reported overall accuracy measures which do not adequately reflect the technique's performance (as will be shown in chapter 5). One major drawback of unsupervised methods is that they often generate clusters that are undesired or have no meaning in the real world. For example, clustering of forum posts on the travel domain might lead to a cluster containing posts pertaining to New York City sightseeing alone. This is a meaningful cluster in general, but it has no meaning when one aims to find clusters of post categories such as \textit{Question}, \textit{Answer}, \textit{Feedback}, etc. Moreover, because the clusters are unlabeled, post-processing is necessary to map the clusters to the categories that are desired as the output.

\section{Semi-supervised Methods}
Semi-supervised methods can overcome the drawbacks of both unsupervised and supervised methods by using a minimal amount of labeled data (that is costly to obtain) and a large amount of unlabeled data (that is easily available). To the best of our knowledge, there exist only two semi-supervised methods for categorization of posts in Web forums. One employed domain adaptation from labeled spoken dialogue datasets by means of a sub-tree pattern mining algorithm \citep{Jeong2009}. Another method extracted \textit{Answer} posts in forum threads using a co-training framework \citep{Catherine2013}. However, it focused only on extracting \textit{Answer} posts, with the assumption that the first post in a thread is a \textit{Question}. Both methods used features such as the chronological position of a post in the thread, and post and author ratings.

\section{Methods Applied to Other Tasks}
There exists other previous work that is applied to tasks unrelated to forum post categorization but which inspires the development of techniques discussed in this research paper. \cite{Barzilay2004} proposed a content model for multi-document summarization based on sentence extraction. This model consists of an HMM at the sentence level that is tailored towards identifying sentence clusters belonging to different topics. Inspired by this model, \cite{Ritter2010} suggested a `conversation model' for the modeling of dialogue acts in Twitter conversations. Their model replicates Barzilay's model but replaces sentences in a document with tweets in a Twitter conversation as units of the HMM. They used Topic Modeling (using Latent Dirichlet Allocation) along with the conversation model and reported better performance; but the evaluation was done only qualitatively. Similarly, \cite{Joty2011} applied conversation models to email and forum threads where a single post is considered an HMM unit. They further enriched this technique by using structural features from emails and forums, in addition to language models. They used GMMs along with their feature-enhanced conversation models, and reported better performance than using conversation models alone. The motivation for these techniques is that HMMs can model the sequential nature of dialogue acts well. For example, the fact that a \textit{Solution} is more likely to follow a \textit{Problem}, as opposed to any other category, can be implicitly encoded in the HMMs.
\chapter{Description of Implemented Methods}
The code for existing methods (that are relevant to this work) is not available to other
researchers. Also, a number of technical details that are necessary
for reproduction are omitted in literature. Hence, it is important to
describe the implementations of previous methods that inspire or form
the basis of the proposed methods. In the process, a few enhancements are
also proposed. These are described in the following section.

\section{Existing Methods with Minor Enhancements}
\subsection{Conversation Model}
\begin{figure}[t!]
\centering
\includegraphics[width=90mm]{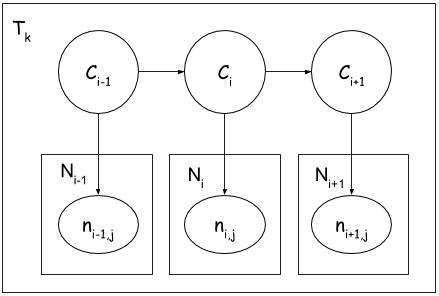}
\caption{Plate notation of conversation model\label{fig:conversation-model-plate-notation}}
\end{figure}

The conversation model that was introduced in the previous
chapter is described here. While there are three different variants of
this model (as described in the previous chapter), this work
implements the originally proposed model by \cite{Barzilay2004},
while making necessary modifications for applying it to forum post
categorization. The conversation model is a Hidden Markov Model (HMM),
in which hidden (unobserved) states correspond to post categories, and
emissions (observed) correspond to bags of post n-grams. A plate notation
of the equivalent graphical model is shown in Figure
\ref{fig:conversation-model-plate-notation} (derived from
\cite{Ritter2010} and \cite{Joty2011}). Here, a thread $T_{k}$
consists of a sequence of category labels, and each category label $C_{i}$ emits a bag of word
n-grams $N_{i}$ of the $i^{\text{th}}$
chronological post in the thread.

The priors for this model are derived from a two-step process: (i) every post is
represented as a vector of word n-gram frequency counts, and (ii) the
vectors are clustered using hierarchical clustering. The resultant
cluster labels are used to calculate the frequency counts of initial
HMM states and state transitions, and hence, the corresponding
probabilities. The priors are optionally calculated using an additional concept of
\textit{insertion states}. These are the states which contain a number of posts
fewer than a fixed threshold, called \textit{state size
  threshold}. This concept is used to account for small noise states
that pertain to no meaningful target category. If used, all insertion
states are merged into a single state, representing a noise
state.

The learning algorithm (Algorithm \ref{algorithm:conversation-model}) of the conversation model uses iterative
Expectation Maximization (EM) to maximize the expected
probability of a post given a state, repeating until convergence of
the sum of all observation probabilities. During the expectation step
(E-step), a word n-gram language model is constructed for each
state. Using this state-specific language model, the emission
probability of an observation (or post) can be calculated. During the maximization step (M-step), the most likely
state sequence is calculated using Viterbi algorithm. All
configuration parameters used in this algorithm are described in
Table \ref{table:config-params}. Each function used in the algorithm is described below.

\begin{itemize}
\item{\textit{vectorize}}: Given a post, it outputs a vector
  using frequency counts of word n-grams in the post. The
  number of dimensions of the vector is equal to the word vocabulary size
  of all posts.
\item{\textit{cluster}}: Given a set of vectors, it clusters them using the complete linkage
  hierarchical clustering algorithm with cosine distance metric, and
  outputs a cluster label for each vector.
\item{\textit{merge\_small\_states}}: Given a list of states (one for
  each post), it merges all states with fewer than
  \textit{stateSizeThreshold} number of posts into a single state, and outputs the updated
  states as well as the updated number of states. This is applicable only if
  the \textit{mergeInsertionStates} parameter is set to \textit{true}.
\item{\textit{language\_model}}:  Constructs a word n-gram language model
  for the posts belonging to a given state. A smoothing parameter
  $\delta_1$ is used to account for unseen word n-grams when calculating
  the probability of a post.
\item{\textit{Viterbi\_algorithm}}: Runs Viterbi algorithm to output
  the most likely state sequence, given the HMM
  parameters (i.e., initial state probabilities, state transition
  probabilities, and state-specific language models).
\end{itemize}

\clearpage

In the HMM, the probability of a post $P_i$, given a state $S_k$, is
calculated as a categorical probability of its word n-grams, as shown in Equation \ref{eq:cm}.

\begin{equation} \label{eq:cm}
p(P_i|S_k) = \prod_j p(W_{i,j}|L_k)
\end{equation}
where:
\begin{itemize}[noitemsep,label=]
  \item $W_{i,j}$ is the $j^{\text{th}}$ (in no particular order) word
    n-gram in post $P_i$,
  \item and $L_k$ is the language model for state $S_k$.
\end{itemize}

\begin{table}[b!]
\centering
{\tabulinesep=2mm
\begin{tabu}{p{4.5cm} p{8cm} p{2.5cm}}
\hline
\textbf{Parameter Name} & \textbf{Description} & \textbf{Data Type} \\
\hline
\textit{initialNumClusters} & The initial number of clusters to be
output using agglomerative clustering & Integer \\
\hline
\textit{mergeInsertionStates}; \hspace{0.5cm}
\textit{stateSizeThreshold} & States with a number of posts fewer than
\textit{stateSizeThreshold} are merged into a single state if
\textit{mergeInsertionStates} is set to \textit{true}
& Boolean; \hspace{0.5cm} Integer \\
\hline
\textit{lmType} & The type of language model to be used for calculating the
emission probability of a post given a state & `unigram'
\hspace{0.5cm} or \hspace{1.5cm}
`bigram' \\
\hline
$\delta_1$ & Smoothing parameter for language modeling (to account for
unseen n-grams) & Float \\
\hline
$\delta_2$ & Smoothing parameter for calculation of HMM state transition
probabilities (to account for unseen state transitions) & Float \\
\hline
\textit{maxNumIterations} & Maximum number of iterations of
Expectation Maximization & Integer \\
\hline
\textit{numMixtureComponents} & Number of mixture components to be used for conversation
model with Gaussian mixtures & Integer \\
\hline
\end{tabu}
}
\caption{\label{table:config-params} Configuration parameters used in conversation models}
\end{table}

\begin{algorithm}[t!]
\algsetup{linenosize=\footnotesize}
\footnotesize
\caption{Conversation model}
\label{algorithm:conversation-model}
\vspace{0.1cm}
\textbf{Input}: A list of threads \textit{T}, each containing a list
of posts \textit{P} (in chronological order)\\
\textbf{Parameters}: \textit{initialNumClusters}, \textit{mergeInsertionStates},
\textit{stateSizeThreshold}, \textit{maxNumIterations},
\textit{lmType}, $\delta_1$, $\delta_2$\\
\textbf{Output}: A list of cluster labels \textit{CL} for each post in
each thread (in the order of the input)
\begin{algorithmic}[1]
\FORALL{$\text{thread } T_x$}
  \FORALL{$\text{post } P_{x,y} \in T_x $}
    \STATE $V_{x,y} := vectorize(P_{x,y})$ \hspace{0.5cm} // $V_{x,y}$ is the vector of post $P_{x,y}$
  \ENDFOR
\ENDFOR
\STATE $ICL := cluster(V, initialNumClusters)$ \hspace{0.5cm} // $ICL$ is
the list of initial cluster labels for each post ($ICL_{x,y}$ is the
initial cluster label for post $P_{x,y}$ in thread $T_x$)
\STATE $S := ICL $ \hspace{0.5cm} // $S$ is the list of states for
all posts; at this step, it is the same as the initial cluster labels
\FOR{$n = 1 \to maxNumIterations$}
  \IF{$mergeInsertionStates$ is $true$}
    \STATE $[S,numStates] := merge\_small\_states(S,stateSizeThreshold)$
  \ENDIF
  \FOR{$i = 1 \to numStates$}
    \STATE $SP_i = \emptyset$
    \FORALL{$\text{state } S_{x,y}$}
      \IF{$S_{x,y} = i$}
        \STATE $SP_i := SP_i \cup P_{x,y}$ \hspace{0.5cm} // $SP_i$ is
        the set of all posts that belong to state $i$
      \ENDIF
    \ENDFOR
    \STATE $L_i := language\_model(SP_i,lmType,\delta_1)$
  \ENDFOR
  \FOR{$i = 1 \to numStates$}
    \STATE $init\_counts_{i} := \Sigma_{T_x} \mathbbm{1}{S_{x,1} = i} $
    \hspace{0.5cm} // $S_{x,1}$ is the state of the first post in thread $T_x$
  \ENDFOR
  \FOR{$i = 1 \to numStates$}
    \STATE $\pi_{i} := (init\_counts_{i} + \delta_2) /(\Sigma_{k} (init\_counts_{k}) +
    \delta_2 \times numStates)$ \hspace{0.3cm}// $\pi_{i}$ is the
    probability that initial state is $i$
  \ENDFOR
  \FOR{$i = 1 \to numStates$}
    \FOR{$j = 1 \to numStates$}
      \STATE $trans\_counts_{i,j} := \sum_{T_x} \sum\limits_{a=1}^{|T_x| - 1} \mathbbm{1}{S_{x,a} = i, S_{x,a+1} = j}$
    \ENDFOR
  \ENDFOR
  \FOR{$i = 1 \to numStates$}
    \FOR{$j = 1 \to numStates$}
      \STATE $\phi_{i,j} := (trans\_counts_{i,j} +
      \delta_2)/(\Sigma_{k,l} (trans\_counts_{k,l}) + \delta_2 \times
      numStates^2)$ \hspace{0.3cm} // $\phi_{i,j}$ is the probability of transitioning from
state $i$ to state $j$
    \ENDFOR
  \ENDFOR
  \STATE $S := Viterbi\_algorithm(\pi, \phi, L)$
  \IF{sum of observation probabilities converged}
    \STATE \textbf{break}
  \ENDIF
\ENDFOR
\STATE $CL := S$
\end{algorithmic}
\end{algorithm}

\clearpage

\subsection{Conversation Model with Gaussian Mixtures}
\begin{figure}[t!]
\centering
\includegraphics[width=90mm]{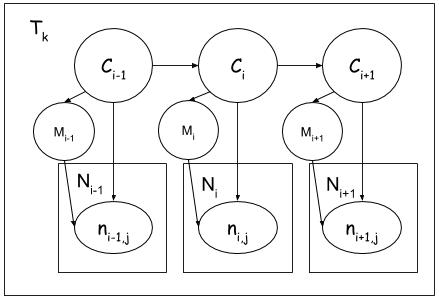}
\caption{Plate notation of conversation model with Gaussian Mixtures\label{fig:mixture-model-plate-notation}}
\end{figure}

The previous model used standard HMM emission probabilities that were
based on n-gram frequency counts, which can suffer from the drawback of
producing topical clusters. To counter this, \cite{Joty2011} proposed a
method which models the HMM emissions as a mixture of Gaussians, i.e.,
a Gaussian Mixture Model (GMM). A plate notation of the resultant model is shown in Figure \ref{fig:mixture-model-plate-notation}. Here a thread $T_{k}$ consists of a
sequence of category labels, and each category label $C_{i}$ and Gaussian mixture $M_i$ emit a bag of
word n-grams $N_i$, which corresponds to the $i^{\text{th}}$ chronological
post in the thread.  Apart from preventing topical clusters, the authors argue
that this can define finer and hence, richer emission
distributions. Also, in contrast to the Topic Model-based approach \citep{Ritter2010},
learning and inference can be done using the EM algorithm without approximate inference techniques.

In addition to the steps in the simple conversation model, the learning algorithm (Algorithm \ref{algorithm:mixture-model}) of
the current model uses Gaussian mixture components as input to the
Viterbi algorithm to calculate the most likely state sequence. Each function used in the algorithm is described below.
\begin{itemize}
\item{$fit\_GMM$} -- It fits the given vector to the GMM corresponding
  to the vector's state. The initial values of mean and
  variance of each mixture component are initialized randomly. The
  value of the \textit{numMixtureComponents} parameter decides the number of mixture components to
  be used.
\item{$Viterbi\_algorithm$} -- Runs Viterbi algorithm in order to
  output the most likely state sequence given the parameters of the
  HMM and GMMs (i.e., initial state probabilities, state transition
  probabilities, and state-specific Gaussian mixture components).
\end{itemize}

Here, the probability of a post $P_i$, given a state $S_k$, is
calculated as shown in
Equation \ref{eq:cmgmm}.
\begin{equation} \label{eq:cmgmm}
p(P_i|S_k) = \sum_j p(M_{k,j}|S_k) p(P_i|M_{k,j})
\end{equation}
where:
\begin{itemize}[noitemsep,label=]
  \item $M_{k,j}$ refers to the $j^{\text{th}}$  (in
    no particular order) mixture model component for state $S_k$.
\end{itemize}

\subsection{Fully Supervised Methods}
Accumulating all features used by existing supervised methods and
modifying them to suit specific datasets (where necessary), a fully
supervised method is implemented using Support Vector Machines
(SVM)\footnote{The weka.classifiers.functions.SMO classifier from the Weka toolkit \citep{Hall2009weka} 
  is used for implementing SVM.}. Table \ref{table:supervised-features} lists the most representative features
that were used.

\begin{algorithm}[t!]
\algsetup{linenosize=\footnotesize}
\footnotesize
\caption{Conversation model with Gaussian Mixtures}
\label{algorithm:mixture-model}
\vspace{0.1cm}
\textbf{Input}: A list of threads \textit{T}, each containing a list
of posts \textit{P} (in chronological order)\\
\textbf{Parameters}: \textit{initialNumClusters}, \textit{mergeInsertionStates},
\textit{stateSizeThreshold}, \textit{maxNumIterations},
\textit{lmType}, $\delta_1$, $\delta_2$, \textit{numMixtureComponents}\\
\textbf{Output}: A list of cluster labels \textit{CL} for each post in
each thread (in the order of the input)
\begin{algorithmic}[1]
\FORALL{$\text{thread } T_x$}
  \FORALL{$\text{post } P_{x,y} \in T_x $}
    \STATE $V_{x,y} := vectorize(P_{x,y})$ \hspace{0.5cm} // $V_{x,y}$ is the vector of post $P_{x,y}$
  \ENDFOR
\ENDFOR
\STATE $ICL := cluster(V, initialNumClusters)$ \hspace{0.5cm} // $ICL$ is
the list of initial cluster labels for each post ($ICL_{x,y}$ is the
initial cluster label for post $P_{x,y}$ in thread $T_x$).
\STATE $S := ICL $ \hspace{0.5cm} // $S$ is the list of states for
all posts; at this step, it is the same as the initial cluster labels.
\FOR{$n = 1 \to maxNumIterations$}
  \IF{$mergeInsertionStates$ is $true$}
    \STATE $[S,numStates] := merge\_small\_states(S,stateSizeThreshold)$
  \ENDIF
  \FORALL{$\text{thread } T_x$}
    \FORALL{$\text{post } P_{x,y} \in T_x $}
      \STATE $fit\_GMM(G_{S_{x,y}},P_{x,y},numMixtureComponents)$
      \hspace{0.3cm} // $G$ is the set of GMMs; $G_i$ is the GMM for state $i$
    \ENDFOR
  \ENDFOR
  \FOR{$i = 1 \to numStates$}
    \STATE $init\_counts_{i} := \Sigma_{T_x} \mathbbm{1}{S_{x,1} = i} $
    \hspace{0.5cm} // $S_{x,1}$ is the state of the first post in thread $T_x$
  \ENDFOR
  \FOR{$i = 1 \to numStates$}
    \STATE $\pi_{i} := (init\_counts_{i} + \delta_2) /(\Sigma_{k} (init\_counts_{k}) +
    \delta_2 \times numStates)$ \hspace{0.3cm}// $\pi_{i}$ is the
    probability that initial state is $i$
  \ENDFOR
  \FOR{$i = 1 \to numStates$}
    \FOR{$j = 1 \to numStates$}
      \STATE $trans\_counts_{i,j} := \sum_{T_x} \sum\limits_{a=1}^{|T_x| - 1} \mathbbm{1}{S_{x,a} = i, S_{x,a+1} = j}$
    \ENDFOR
  \ENDFOR
  \FOR{$i = 1 \to numStates$}
    \FOR{$j = 1 \to numStates$}
      \STATE $\phi_{i,j} := (trans\_counts_{i,j} +
      \delta_2)/(\Sigma_{k,l} (trans\_counts_{k,l}) + \delta_2 \times
      numStates^2)$ \hspace{0.3cm} // $\phi_{i,j}$ is the probability of transitioning from
state $i$ to state $j$
    \ENDFOR
  \ENDFOR
  \STATE $S := Viterbi\_algorithm(\pi, \phi, G)$
  \IF{sum of observation probabilities converged}
    \STATE \textbf{break}
  \ENDIF
\ENDFOR
\STATE $CL := S$
\end{algorithmic}
\end{algorithm}
\clearpage

\clearpage

\begin{table}[t!]
\centering
\resizebox{\columnwidth}{!}{
{\tabulinesep=1mm
\begin{tabu}{>{\quad} l l}
\toprule
\textbf{Feature} & \textbf{Type} \\
\midrule
\rowgroup{\textit{Structural}} \\
Chronological position of post in thread & Numeric  \\
Number of posts in thread & Numeric \\
\midrule
\rowgroup{\textit{Metadata}} \\
Total number of posts in the thread by author of current post & Numeric \\
Total number of previous posts in thread by author of current post & Numeric \\
\midrule
\rowgroup{\textit{Textual}} \\
Average similarity of post to other posts in thread & Numeric \\
Similarity of post to initial post & Numeric \\
Word bigrams & Binary \\
POS bigrams and trigrams & Binary \\
\midrule
\rowgroup{\textit{Language}} \\
Number of tokens in post after stopword removal & Numeric \\
Number of unique tokens after stopword removal & Numeric  \\
Number of unique tokens after stopword removal and stemming & Numeric\\
Number of adverbs & Numeric\\
Number of modal verbs & Numeric\\
Number of nouns & Numeric\\
Number of proper nouns & Numeric\\
Number of \textit{wh}-words (\textit{why}, \textit{where}, \textit{what}, \textit{when}, \textit{how}) & Numeric\\
Number of determiners & Numeric \\
Number of stopwords & Numeric \\
Presence of periods & Binary \\
Presence of question marks & Binary \\
Presence of other punctuation marks & Binary \\
Presence of token -- \textit{thanks} & Binary \\
Presence of token -- \textit{same} or \textit{similar} & Binary \\
Presence of token -- \textit{did} & Binary \\
\bottomrule
\end{tabu}
}
}
\caption{\label{table:supervised-features} Most representative
  features used in implementation of existing fully supervised method.}
\end{table}

\clearpage

\section{Proposed Methods}
\subsection{Conversation Model with Part-of-Speech Tags}
Since conversation models take only word n-gram language models into
account, it is likely that they output clusters of posts that are
topically related, without reflecting the posts' purpose or intention. To
overcome this limitation, the conversation model is enhanced by modeling
emissions as arising partially from part-of-speech (POS) tags of
words. This might better characterize the syntactic nature of the
post. This is based on the assumption that posts belonging to the same
category are likely to be syntactically similar. The proposed model
uses POS n-gram language models in addition to word n-gram language
models, and calculates the HMM emission probability of a post given
its state using a linear combination of both. Here, the probability of
a post $P_i$, given a state $S_k$, is calculated as shown in Equation \ref{eq:cmp}.
\begin{equation} \label{eq:cmp}
\begin{split}
p(P_i|S_k) = \frac{\prod_j \left[ \lambda \times p(W_{i,j}|L_k) + (1 -
  \lambda) \times p(POS_{i,j}|PL_k) \right]}{Z} \\
0\leq\lambda\leq1 \\
Z = \sum_{i,k} \left[ \prod_j \left[ \lambda \times p(W_{i,j}|L_k) + (1 - \lambda)
\times p(POS_{i,j}|PL_k) \right] \right]
\end{split}
\end{equation}
where:
\begin{itemize}[noitemsep,label=]
  \item $POS_{i,j}$ is the $j^{\text{th}}$ (in no particular order) POS n-gram in post $P_i$,
  \item $PL_k$ is the POS n-gram language model for state $S_k$,
  \item $\lambda$ is the parameter that controls the proportion of
    probability arising from the word and POS language models (using
    $\lambda = 1$ is equivalent to the conversation model),
  \item and $Z$ is the normalizing constant.
\end{itemize}

\clearpage

\begin{table}[t!]
\centering
{\tabulinesep=2mm
\begin{tabu}{>{\quad} l l}
\toprule
\textbf{Feature} & \textbf{Type} \\
\midrule
\rowgroup{\textit{Structural}} \\
Chronological position of post in thread & Numeric \\
\midrule
\rowgroup{\textit{Metadata}} \\
Identity of author of current post & Nominal \\
Previous post by same author & Binary \\
Total number of posts in the thread by author of current post & Numeric \\
Total number of previous posts in thread by author of current post & Numeric \\
\midrule
\rowgroup{\textit{Textual}} \\
Number of tokens & Numeric \\
Type to token ratio & Numeric \\
Average similarity of post to other posts in thread & Numeric \\
Similarity of post to initial post & Numeric \\
\midrule
\rowgroup{\textit{Language}} \\
Presence of question marks & Binary \\
Presence of question marks in previous post & Binary \\
Presence of exclamation marks & Binary \\
Presence of Quotes/URLs/Images & Binary \\
Presence of token -- \textit{thanks} & Binary \\
Presence of token -- \textit{same} or \textit{similar} & Binary \\
Presence of token -- \textit{did} & Binary \\
Number of \textit{wh}-words (\textit{why}, \textit{where}, \textit{what}, \textit{when}, \textit{how}) & Numeric\\
Number of modal verbs & Numeric \\
Number of proper nouns & Numeric \\
\bottomrule
\end{tabu}
}
\caption{\label{table:semi-supervised-features} All features
used in proposed methods with feature models.}
\end{table}

\clearpage

\subsection{Conversation Model with Features}
This model allows for the incorporation of discriminative features
that might be useful for generating clusters that better represent the
desired categories. For example, the chronological position of a post
in a thread might be a useful feature, because a post is more likely
to be a \textit{Problem} if it is the first post in a thread as opposed to any
other position. Here, the probability of a post $P_i$, given a state
$S_k$, is calculated as shown in equation \ref{eq:cmf}.

\begin{equation} \label{eq:cmf}
p(P_i|S_k) = \prod_j p(W_{i,j}|L_k) \prod_f p(F_{i,f}|FL_k)
\end{equation}
where:
\begin{itemize}[noitemsep,label=]
  \item $F_{i,f}$ is the $f^{\text{th}}$ (in no particular
    order) discrete-valued feature in post $P_i$,
  \item and $FL_k$ is the feature model for state $S_k$.
\end{itemize}

Table \ref{table:semi-supervised-features} lists the features used in
this model. All feature values are discretized. These features
comprise a small subset of those used in the fully supervised
setup, and are relatively simpler and easier to obtain.

\subsection{Conversation Model with Post Embeddings}
In the conversation models, the clustering of posts is performed as a
first step using vectors of word n-grams in the post. This step may
suffer from issues of sparsity and high vector dimensionality. To
avoid this, it is proposed to use embeddings that are low-dimensional
semantic representations of
posts. \textit{Word2Vec}\footnote{\url{http://code.google.com/p/word2vec/}}, with
enhancements as proposed by \cite{Le2014}, can be
used to generate embeddings of variable lengths of text. This
technique uses a recurrent neural network that predicts a word given
its surrounding context. For the current task, this technique is used
to generate one embedding per post, which can then be used for clustering. The rest of the model remains unchanged.

\subsection{Semi-supervised Conversation Model}
As discussed before, semi-supervised techniques can make use of
a minimal amount of labeled data in order to better guide the prediction
of labels (as opposed to unlabeled clusters in case of unsupervised
techniques). A modification can be made to the previous models
to achieve this --- the priors can be constructed from a small amount of labeled data instead of
clustering all posts using vectors of post n-grams. More concretely,
labeled data can be used to initialize the language models and the HMM
parameters (initial state and state transition probabilities) for the
first iteration of the EM algorithm. The rest of the model
remains unaffected.

\subsection{Other Enhancements}
All the models discussed above can be combined with one another, except
in the case of semi-supervised models with post embeddings. This is because the semi-supervised
models calculate priors from labeled data, whereas those with post
embeddings use hierarchical clustering of unlabeled data.

Also, the following modifications can be made in an attempt to
simplify the models and improve performance. The conversation
models with POS tags require the setting of a configuration parameter
which decides the proportion of probability that comes from language
and POS models in the linear combination. Also, this parameter value
(when fixed) is used uniformly across all word and POS n-grams. However,
one could estimate a parameter value that is specific to a word and
POS tag pair by using frequency counts from predicted labels during the previous iteration of the EM
algorithm. In case of the first iteration of the unsupervised models, the frequency counts can be
calculated using the initial cluster labels; and in case of
semi-supervised models, this can be done using the labels of the training
data. Equation \ref{eq:pos-frac1} can be used to calculate the
fractional contribution of a word in the language model $L_{k}$ for
state $S_k$, and equation \ref{eq:pos-frac2} can be used analogously
for calculating the fractional contribution of a POS tag. Equation \ref{eq:pos-frac3} can be used to determine
the value of $\lambda$, which can then be used in the conversation model
with POS tags, as shown in equation \ref{eq:cmp-pos-frac}.

Discussion forum posts often contain informal text with misspellings
and spelling variations, which cannot be modeled by word n-gram
language modeling. However, character n-grams could potentially
overcome this limitation. Also, they have been a very useful
discriminative feature in the area of authorship attribution, because
they seem to account for lexical, syntactic, and stylistic information
\citep{Sapkota2015}. Hence, character n-gram language models can be
used in isolation or in addition to word n-gram language models in
each of the models discussed in previous sub-sections.

\begin{equation} \label{eq:pos-frac1}
WordFrac(L_{k},w) = \frac{\text{Frequency of }w \text{ in posts from state }S_k}{\text{Total frequency of }w}
\end{equation}

\begin{equation} \label{eq:pos-frac2}
PosFrac(PL_{k},pos) = \frac{\text{Frequency of }pos \text{ in posts
    from state }S_k}{\text{Total frequency of } pos}
\end{equation}

\begin{equation} \label{eq:pos-frac3}
\lambda (w,pos,k) = \frac{WordFrac(L_{k},w)}{WordFrac(L_{k},w) + PosFrac(PL_{k},pos)}
\end{equation}

\begin{equation} \label{eq:cmp-pos-frac}
\begin{split}
p(P_i|S_k) = \frac{\prod_j \left[ \lambda (W_{i,j},POS_{i,j},k) \times
p(W_{i,j}|L_k) + (1 - \lambda (W_{i,j},POS_{i,j},k)) \times
p(POS_{i,j}|PL_k) \right]}{Z} \\\\
Z = \sum_{i,k} \left[ \prod_j \left[ \lambda (W_{i,j},POS_{i,j},k) \times
p(W_{i,j}|L_k) + (1 - \lambda (W_{i,j},POS_{i,j},k)) \times
p(POS_{i,j}|PL_k) \right] \right]
\end{split}
\end{equation}

\section{Mapping of Clusters to Categories}
Unsupervised methods output cluster labels for each post (and not a
specific category label). In order to match them with an observed
category label, a one-to-one mapping is obtained using Kuhn-Munkres
algorithm for maximal weighting in a bipartite graph
\citep{Kuhn1955,Munkres1957}. In this procedure, one set of disjoint nodes of the
bipartite graph corresponds to the set of predicted cluster labels, and
the other set corresponds to the set of manually obtained gold
labels. The weight of an edge from cluster label $c$ to gold label $g$
is calculated as the number of posts which are predicted as $c$ and
also have a gold label $g$. \cite{Joty2011} follow the same procedure.
\chapter{Data Collection and Annotation}
Previous work has used forum datasets belonging to the travel and
computer-related technical domains (listed in Table
\ref{table:existing-datasets}).

\begin{table}[b!]
\centering
\resizebox{\columnwidth}{!}{
{\tabulinesep=1mm
\begin{tabu}{p{15cm}}
\midrule
\textbf{Ubuntu} \citep{Bhatia2012}\\
\hspace{1em}Domain: Computer technical\\
\hspace{1em}Tagset: \textit{Question}, \textit{Repeat Question}, \textit{Clarification}, \textit{Solution}, \textit{Further Details}, \\
\hangpara{2em}{0}\textit{Positive Feedback}, \textit{Negative Feedback}, \textit{Spam}\\
\hspace{1em}Number of threads: 100\\
\midrule
\textbf{TripAdvisor-NYC} \citep{Bhatia2012}\\
\hspace{1em}Domain: Travel\\
\hspace{1em}Tagset: Same as \textbf{Ubuntu}
\hspace{1em}Number of threads: 100\\
\midrule
\textbf{Apple} \citep{Catherine2012}\\
\hspace{1em}Domain: Computer technical\\
\hspace{1em}Tagset: \textit{Answer}\\
\hspace{1em}Number of threads: 300 labeled and 140,000 unlabeled\\
\bottomrule
\end{tabu}
}
}
\caption{\label{table:existing-datasets} Existing discussion forum
  datasets used in this research paper.}
\end{table}

\begin{table}[t]
\centering
\resizebox{\columnwidth}{!}{
{\tabulinesep=1mm
\begin{tabu}{l p{10cm}}
\hline
\textbf{Post Category} & \textbf{Description} \\
\hline
\textit{Problem} & A query on a particular topic \\
\textit{Solution} & A suggested solution or answer to one of the
previous posts annotated as \textit{Problem} \\
\textit{Clarification-Request} & A query regarding one of the previous
posts annotated as \textit{Problem} or \textit{Solution} \\
\textit{Clarification} & A suggested solution or answer to one of the
previous posts annotated as \textit{Clarification-Request} \\
\textit{Feedback} & A comment about one of the previous posts by a
different user that is annotated as \textit{Solution} \\
\textit{Other} & The post does not belong to any of the previous
categories \\
\hline
\end{tabu}
}
}
\caption{\label{table:post-category-tagset} Tagset of forum post
  categories used for annotating the Verticalscope datasets.}
\end{table}

In addition to these, the current work attempts to
observe the performance of post categorization on forums belonging to the
automotive domain. For this purpose, forums that discuss Jeep and
Mercedes-Benz vehicles were obtained from Verticalscope
Inc.\footnote{Verticalscope Inc.\@ (http://www.verticalscope.com) is a
  privately held corporation that specializes in the acquisition and
  development of websites and online communities for the Automotive,
  Powersports, Power Equipment, Pets, Sports and Technology vertical
  markets.} Around 150 threads each were randomly picked from
JeepForum\footnote{\url{jeepforum.com}} and
BenzWorld\footnote{\url{benzworld.org}}. Threads whose first posts
contained advertisements or spam posts (as identified by Topic
Modeling done previously) were filtered out. Also, threads which had
only one post or more than 30 posts, were discarded. This resulted in
a total of 93 threads in the JeepForum dataset, and 108 threads
in the Benzworld dataset.

Next, previous literature was studied in order to decide
the tagset of categories to annotate the forum posts in the dataset. \cite{Kim2010} use a tagset of 12 categories --- \textit{Question},
\textit{Question-Add}, \textit{Question-Confirmation},
\textit{Question-Correction}, \textit{Answer}, \textit{Answer-Add},
\textit{Answer-Confirmation}, \textit{Answer-Correction},
\textit{Answer-Objection}, \textit{Resolution}, \textit{Reproduction},
and \textit{Other}. Since this is the most
fine-grained set of categories, a pilot annotation study
was conducted using these. Five annotators annotated posts from six randomly picked threads
in the automotive domain. Based on the quantitative results
of the annotation and the feedback from annotators, it was observed that using
a more coarse-grained set of six categories would be simpler and more
meaningful. These categories and their description are shown in Table \ref{table:post-category-tagset}.

The main annotation task was set up on CrowdFlower\footnote{\url{http://www.crowdflower.com/}}, a Web
platform for obtaining crowdsourced annotations. In each thread, the
posts were displayed to the annotators in chronological order. Some posts contain
quoted text, i.e., a span of text from a previously posted
answer. These were enclosed within `[QUOTE]' tags along with the
username of the post which is quoted. Some posts contain URLs or
images, which were displayed to the annotators using the tags `[URL]' and `[IMG]'
respectively. In addition to providing the target set of annotation categories, the following instructions were provided to the annotators.
\vspace{-0.5em}
\begin{itemize}[noitemsep]
\item Every post must have exactly one category associated with it. If there is confusion between multiple categories for a single post, choose the category that describes the main purpose of the post.
\item \textit{Clarification-Request} is also a type of post that
  discusses a problem, but it must relate to an earlier post annotated as \textit{Problem} or \textit{Solution}.
\item \textit{Clarification} is also a type of post discussing a solution, but it must relate to an earlier post annotated as \textit{Clarification-Request}.
\end{itemize}

\vspace{1cm}

\begin{table}[h!]
\centering
\resizebox{\columnwidth}{!}{
{\tabulinesep=2mm
\begin{tabu}{c c c c}
\hline
\textbf{Forum} & \textbf{\# Threads} & \textbf{\% Majority Annotations} & \textbf{Krippendorf's $\alpha$} \\
\hline
JeepForum & 93 & 93\% & 0.62 \\
BenzWorld & 108 & 77\% & 0.47 \\
\hline
\end{tabu}
}
}
\caption{\label{table:verticalscope-datasets} The number of threads in
  the Verticalscope datasets, along with measures of the quality of
  forum post annotations (i.e., the percentages of majority annotations and inter-annotator
  agreement values).}
\end{table}

\clearpage

\begin{figure}[t!]
\centering
\includegraphics[width=90mm]{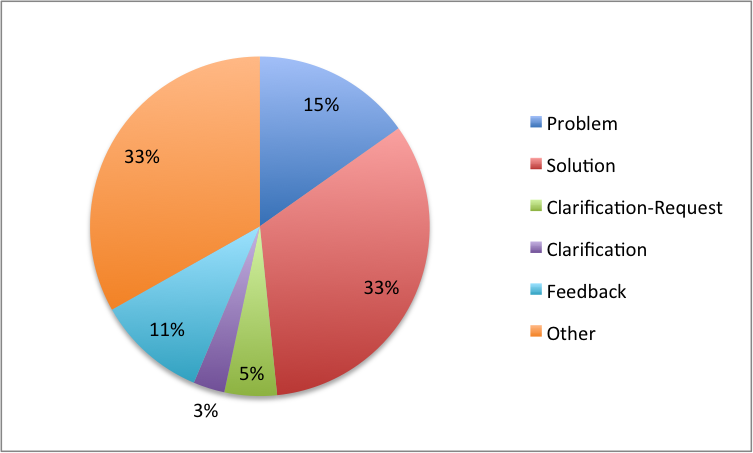}
\caption{Category-wise distribution of posts in the JeepForum dataset. \label{fig:jeepforum-label-distribution}}
\end{figure}

\begin{figure}[t!]
\centering
\includegraphics[width=90mm]{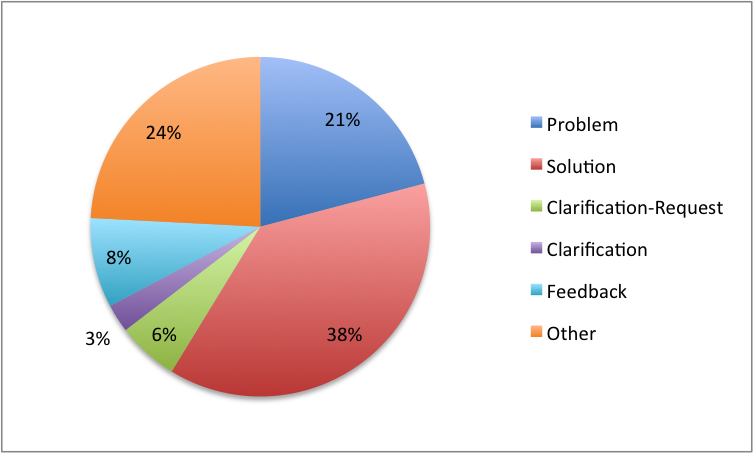}
\caption{Category-wise distribution of posts in the BenzWorld dataset. \label{fig:benzworld-label-distribution}}
\end{figure}

The top three trusted annotators were picked from each annotation task, and
gold labels were assigned to each post if at least two out of three
annotators agreed. However, if there was disagreement among all
annotators, the post was left unlabeled. Details of the resulting
datasets, including the inter-annotator agreements and quantity of
data, are shown in Table \ref{table:verticalscope-datasets}. Since
the annotations were crowdsourced, there is no common set of
annotators for each post. Hence, instead of using standard annotation
quality measures like Scott's $\pi$ and Fleiss's $\kappa$,
Krippendorf's $\alpha$ is reported, which can account for missing
values \citep{Artstein2008}. The values obtained (i.e., 0.62 and 0.47)
are reflective of moderate to substantial agreement. In order to
further confirm the validity of the annotations, two annotators
randomly sampled 10\% of the threads and manually analyzed the
annotations for correctness. They found 98.8\% and 91.8\% of posts to
be correctly annotated in the JeepForum and BenzWorld datasets respectively.

The category-wise distribution of posts for both datasets are shown in Figures \ref{fig:jeepforum-label-distribution} and \ref{fig:benzworld-label-distribution}. \textit{Solution} and \textit{Other} are the most prevalent categories, whereas \textit{Clarification-Request} and \textit{Clarification} form only 8-9\% of the posts. Consequently, the former two categories are expected to be easier to classify (i.e., achieve better accuracy in classification) in comparison to the latter two.

\chapter{Experiments}
\section{Evaluation Measures}
The predicted labels for all posts can be evaluated against the
corresponding gold labels using metrics like precision, recall and
$F_{1}$-measure. Moreover, micro-averaged and macro-averaged
values of these metrics can indicate overall performance across
categories. All evaluation metrics are calculated as shown in
Equations \ref{eq:accuracy} to \ref{eq:macro-f1-measure}. In all cases, \textit{c} is a single
category, and \textit{CS} is the set of all categories. The
values of micro-averaged precision, recall and $F_1$-measure are all
equal if the number of predictions is the same as the number of posts (i.e., every
post is predicted as belonging to some category). All methods
implemented in the current work make some category prediction for every post;
hence, this condition holds true.
\vspace{-0.5em}
\begingroup
\small
\begin{equation} \label{eq:accuracy}
\text{Accuracy, } A(\textit{c}) = \frac{\text{\# actual \textit{c} posts predicted as \textit{c}}
  + \text{\# actual non-\textit{c} posts predicted as non-\textit{c}}}{\text{\#
    predictions}}
\end{equation}
\begin{equation} \label{eq:precision}
\text{Precision, } P(\textit{c}) = \frac{\text{\# actual \textit{c} posts predicted as \textit{c}}}{\text{\# posts predicted as \textit{c}}}
\end{equation}
\begin{equation} \label{eq:recall}
\text{Recall, } R(\textit{c}) = \frac{\text{\# \textit{c} posts predicted as \textit{c}}}{\text{\# actual \textit{c} posts}}
\end{equation}
\begin{equation} \label{eq:f1-measure}
\text{$F_{1}$-Measure, } F(\textit{c}) = \frac{2 \times P \times R}{P + R}
\end{equation}
\begin{equation} \label{eq:micro-accuracy}
\text{Micro-Averaged-Accuracy, } MicroA = \frac{\Sigma_{\textit{c} \in
  \textit{CS}} \left[ \text{\# actual \textit{c} posts predicted as \textit{c}}
\right]}{\text{\# predictions}}
\end{equation}
\begin{equation} \label{eq:micro-precision}
\text{Micro-Averaged-Precision, } MicroP = MicroA
\end{equation}
\begin{equation} \label{eq:micro-recall}
\text{Micro-Averaged-Recall, } MicroR = \frac{\Sigma_{\textit{c} \in
    \textit{CS}} \left[ \text{\# \textit{c} posts predicted as
      \textit{c}} \right]}{\text{\# posts}}
\end{equation}
\begin{equation} \label{eq:micro-f1-measure}
\text{Micro-Averaged-$F_1$-Measure, } MicroF = \frac{2 \times MicroP \times MicroR}{MicroP + MicroR}
\end{equation}
\begin{equation} \label{eq:macro-accuracy}
\text{Macro-Averaged-Accuracy, } MacroA = \frac{\Sigma_{\textit{c} \in
    \textit{CS}} \left[ Accuracy(\textit{c}) \right]}{|\textit{CS}|}
\end{equation}
\begin{equation} \label{eq:macro-precision}
\text{Macro-Averaged-Precision, } MacroP = \frac{\Sigma_{\textit{c}
    \in \textit{CS}} \left[ Precision(\textit{c}) \right]}{|\textit{CS}|}
\end{equation}
\begin{equation} \label{eq:macro-recall}
\text{Macro-Averaged-Recall, } MacroR = \frac{\Sigma_{\textit{c} \in
    \textit{CS}} \left[ Recall(\textit{c}) \right]}{|\textit{CS}|}
\end{equation}
\begin{equation} \label{eq:macro-f1-measure}
\text{Macro-Averaged-$F_1$-Measure, } MacroF = \frac{2 \times MacroP \times MacroR}{MacroP + MacroR}
\end{equation}
\endgroup

\section{Experimental Setup}
\subsection{Preprocessing and Configuration Parameters}
Initially, all forum posts were tokenized by sentence and word,
followed by POS tagging and stemming --- all using Stanford CoreNLP Toolkit
\citep{Manning2014}. Stopword removal was found to degrade
performance; hence, it was not used. It is important to note that
forum conversations often consist of informal English language text, along with the use of domain-specific abbreviations, and
non-standard special characters, such as ellipses and emoticons. Hence,
some errors are introduced in all the previous steps. However, no
effort was made to overcome them, and this is accepted as a limitation
of the current work. 

All methods, except those using post embeddings,
require the conversion of posts to vectors of n-grams. For this
purpose, both unigrams and bigrams were tried, and the former was found
to produce better performance. The use of TF-IDF term weighting did
not improve performance; hence, it was ignored. The maximum number of
iterations of Expectation Maximization was set to 100, which was
sufficient because all experimental runs were completed in fewer than
100 iterations. The values of both smoothing parameters (i.e., \textit{delta1} and \textit{delta2})
were varied in the range of $10^{-1}$ to $10^{-9}$. Subsequently, $10^{-2}$ and
$10^{-9}$ were found to be the best values for \textit{delta1} and
\textit{delta2} respectively. The value of the POS model's $\lambda$ was
varied between $10^{-6}$ and $1 - 10^{-6}$, and the value of 0.999 was found to be the
best. Since the unigram/bigram vocabulary size is much larger than the
POS tag vocabulary size, the former probability distribution is much more
fine-grained. For example, each word unigram's probability value in
the Benzworld dataset is of the order of $10^{-4}$ (since the unigram
vocabulary size is ~5000), whereas each POS unigram's probability
value is of the order of $10^{-2}$ (since the POS vocabulary size is
42). So, the value of 0.999 for word unigrams and 0.001 for POS
unigrams can be viewed as a scaling factor to ensure that both
contribute almost equally towards discriminating between post
categories. To provide further clarity, using a $\lambda$ value of 0.5
gives rise to a predominantly POS-based model because unigram
probability values are too low to make a significant difference
towards identifying one category over another. The parameters, \textit{initialNumClusters} and
\textit{stateSizeThreshold}, directly affect the resulting number
of clusters. In all experimental runs, both these parameters were varied in
the range of 1 to 100, and those which did not output the desired number of
clusters (i.e., number of distinct gold labels) were ignored. In each
case, different parameter values were best suited; however, only the best
performing results are reported. For GMM-based methods, the number of
Gaussian mixture components was varied from 2 to 8, and 3 was found to
be the best value. Parameters specific to GMM, such as initial mixture
component means and variances, were initialized randomly by sampling
from the Gaussian distribution.

For semi-supervised methods, experiments were carried out in a
randomized \textit{n}-fold cross-validation setup. The dataset was
randomly (by sampling from the uniform
distribution) divided into \textit{n} equal-sized folds, and the experiment was run \textit{n} times. In each run, one fold was used for initializing the
priors of the models, and the remaining $\textit{n} - 1$
folds were used for evaluation. This is in contrast to a traditional fully supervised setting,
where $\textit{n} - 1$ folds are used for training and the remaining
fold is used for evaluation.

In the case of language models, it was observed that accuracy values
differ by more than two percentage points when using unigram and
bigram language models. Also, different datasets benefited from
different models. Hence, experiments were run using both, and results
are reported for the better performing alternative.

A number of enhancements were proposed in section 3.2.5 with the
objective of further enhancing the performance of the conversation models.
However, in all cases, these led to deteriorating performance.
Specifically, the use of character or skip-gram language models in isolation or in conjunction
with word language and POS models lowered performance by around 2
percentage points with respect to the best performing method. The use of fractional contributions of language and
POS modeling led to performance deterioration of up to 10 percentage
points. Hence, these enhancements are ignored when reporting results.

\subsection{Baselines}
The \textit{random baseline} randomly (by sampling from the uniform
distribution) assigns category labels to every
post. The \textit{majority
baseline} assigns the most commonly occurring gold category label to
every post. In all datasets on which results are reported, \textit{Solution}
is the most commonly occurring gold category.

Two other baselines are heuristic in nature, and are both based on the assumption that the
first post in the thread is very likely to be a \textit{Problem}. The
first of these, called \textit{Problem-Solution Heuristic
  1}, assigns \textit{Problem} to the first post in the thread, \textit{Other} to the last post, and \textit{Solution}
to the rest. It assumes that the last post in the thread is very
likely to be unrelated to the main thread topic and that many of the
preceding posts are likely to be \textit{Solution}. The second
heuristic baseline, called \textit{Problem-Solution Heuristic
  2}, assigns \textit{Problem} to the first post in the thread, \textit{Solution} to the second post, and
\textit{Other} to the rest. It assumes that the second post is very
likely to be a \textit{Solution} in direct response to the first
\textit{Problem} post, and many of the following posts are likely to be \textit{Other}.

\begin{table}[b!]
\small
\centering
{\tabulinesep=1.75mm
\begin{tabu}{>{\quad} l c c c c}
\cline{2-5}
& \multicolumn{2}{c}{JeepForum} & \multicolumn{2}{c}{BenzWorld} \\
\midrule
\textbf{Model} & \textbf{Micro-A} & \textbf{Macro-A} & \textbf{Micro-A} & \textbf{Macro-A}  \\
\midrule
\rowgroup{\textit{Baselines}} \\
Random & 0.14 & 0.71 & 0.15 & 0.72 \\
Majority & 0.33 & 0.78 & 0.38 & 0.79 \\
Problem-Solution Heuristic 1 & 0.43 & 0.81 & 0.50 & 0.83 \\
Problem-Solution Heuristic 2 & 0.45 & 0.82 & 0.45 & 0.82 \\
\midrule
\rowgroup{\textit{Unsupervised}} \\
CONV & 0.33 & 0.78 & 0.34 & 0.78 \\

CONV + EMB & 0.37 & 0.79 & 0.27 & 0.76 \\

CONV + POS & 0.33 & 0.78 & 0.34 & 0.78 \\

CONV + FEAT & 0.33 & 0.78 & 0.33 & 0.78 \\

CONV + EMB + POS & 0.29 & 0.76 & 0.27 & 0.76 \\

CONV + EMB + FEAT & 0.34 & 0.78 & 0.27 & 0.76 \\

CONV + POS + FEAT & 0.29 & 0.76 & 0.33 & 0.78 \\

CONV + EMB + POS + FEAT & 0.34 & 0.78 & 0.29 & 0.76 \\

CONV + GMM & 0.32 & 0.77 & 0.27 & 0.78 \\ 

CONV + GMM + FEAT & 0.29 & 0.76 & 0.35 & 0.78 \\ 
\midrule
\rowgroup{\textit{Semi-Supervised}} \\
CONV & 0.44 & 0.81 & 0.48 & 0.83 \\ 
CONV + GMM & 0.27 & 0.76 & 0.29 & 0.76 \\
CONV + GMM + FEAT & 0.29 & 0.76 & 0.34 & 0.78 \\
CONV + POS & \textbf{0.48} & \textbf{0.83} & 0.48 & 0.83 \\ 
CONV + FEAT & \textbf{0.49} & \textbf{0.83} & \textbf{0.52} & \textbf{0.84} \\ 
CONV + POS + FEAT & \textbf{0.54} & \textbf{0.85} & \textbf{0.52} & \textbf{0.84}  \\
\bottomrule
\end{tabu}
}
\caption{\label{table:summary-results} Experimental results using
 all the possible combinations of models in both unsupervised and semi-supervised
  settings (CONV: Conversation model; EMB: Post
  embeddings; POS: Part-of-speech model; FEAT: Feature model; GMM: Gaussian
  mixture model). Boldface indicates values that outperform all baselines.}
\end{table}

\clearpage

\section{Main Results}
Table \ref{table:summary-results} lists the micro and
macro-averaged accuracy values when experiments were run using all
possible combinations of the implemented models.

For reported results of unsupervised methods, different values of parameters, \textit{initialNumClusters} and \textit{stateSizeThreshold}, were used in
each case. This is because the same values did not lead to the desired
number of clusters. For example, for the JeepForum dataset,
the conversation model's parameters were: \textit{initialNumClusters =
30} and \textit{stateSizeThreshold = 25}. This resulted in six
clusters, the same as the number of gold label categories. However,
the same parameters yielded a very large number of clusters (15) when used
with the conversation model with post embeddings. Only the best
performing results are reported. In case of methods using GMM, since
parameters were randomly initialized, fluctuations in performance are
expected across different runs. Hence the reported accuracy values
are averages over 10 runs. For the JeepForum dataset, unsupervised
methods reached maximum micro-averaged and macro-averaged accuracy values using
conversation models with post embeddings, POS tags, and
features. However, for the BenzWorld dataset, the performance was the
best using conversation models with GMM and features. All unsupervised
methods outperformed the random baseline. But they performed worse than the
majority baseline in many cases, and the problem-solution heuristic
baselines in all cases.

For semi-supervised methods, the reported accuracy values are averages over 10
runs of 5-fold cross-validation. This setup entails the use of only around 20 labeled
threads for setting the model priors, because both datasets contain
approximately 100 threads. The GMM-based semi-supervised
methods performed only as well as their unsupervised
counterparts. For the JeepForum dataset, semi-supervised methods which used POS tags
and/or features in the absence of GMM, outperformed all baselines. For
the BenzWorld dataset, the same is true, except in case of the
conversation model with POS tags, which performed worse than 
\textit{problem-solution heuristic 1}. Overall, the methods using both POS tags and features
performed the best. For the JeepForum dataset, the best micro-averaged and
macro-averaged accuracy values are 0.54 and 0.85 respectively. In case of
the BenzWorld dataset, the same accuracy values are 0.52 and 0.84
respectively. 

\section{Performance Comparison with State-of-the-Art}
\subsection{Unsupervised HMM+Mix Model}
\cite{Joty2011} reported results of their best performing HMM+Mix
model for dialogue act classification on email and forum thread
datasets, neither of which are available to other researchers. Their forum thread dataset
contains 200 threads sourced from TripAdvisor (for which they report a
macro-accuracy value of 78.35\%). Hence, for performance comparison, the
current work also used a dataset of nearly 200 threads from TripAdvisor (made
available by \cite{Bhatia2012}). As a caveat, it is important to note
that this dataset has eight dialogue act categories, whereas
\cite{Joty2011} consider 12. Also, the current work's conversation model
with GMM (called HMM+Mix++) was used for performance
comparison, since it is an improved adaptation of the HMM+Mix
model. Table \ref{table:comparison-joty-2011} shows that the proposed
conversation model with POS tags and features outperformed
\textit{HMM$+$Mix++} in terms of macro-accuracy values. Also, the
semi-supervised conversation model with POS tags and features performed
much better (0.92 on NYC and 0.90 on Ubuntu); but this is
not directly comparable since the other methods are unsupervised.

\begin{table}[t!]
\centering
{\tabulinesep=2mm
\begin{tabu}{r c  c}
\cline{2-3}
& \textbf{NYC} & \textbf{Ubuntu} \\
\midrule
HMM+Mix++ & 0.85 & 0.83 \\ 
Unsupervised CONV + POS + FEAT & 0.88 & 0.88 \\ 
\bottomrule
\end{tabu}
}
\caption{\label{table:comparison-joty-2011} Experimental results
  comparing the performance of the HMM+Mix++ model with the best
  proposed unsupervised method (i.e., conversation model with POS tags and features).}
\end{table}

\subsection{Semi-supervised Answer Extraction}
\cite{Catherine2013} reported the performance of their semi-supervised
answer extraction approach on 300 labeled threads of the Apple discussion
forums dataset. They trained using only three training threads; however,
these three are not available to other researchers. The code is also
unavailable. Hence, for the sake of simplicity, the methods are
indirectly compared as follows. For their method, values reported in their
paper are used as is. For the best proposed method (i.e., the semi-supervised conversation models with POS tags and features), a 100-fold
cross-validation setup was used (i.e., out of 300 labeled threads, 3
were used for training, and 297
were used for testing, in each fold). Table \ref{table:comparison-catherine-2013} shows that
the values obtained for the proposed method are better in terms of
$F_1$-measure and precision.

\begin{table}[h!]
\centering
{\tabulinesep=2mm
\begin{tabu}{r c  c  c}
\cline{2-4}
& \textbf{Precision} & \textbf{Recall} & \textbf{$F_1$-measure} \\
\hline
\cite{Catherine2013} & 0.57 & 0.84 & 0.68 \\ 
Semi-supervised CONV + POS + FEAT &  0.66 & 0.73 & 0.69 \\
\hline
\end{tabu}
}
\caption{\label{table:comparison-catherine-2013} Experimental results
  comparing the performance of an existing semi-supervised answer
  extraction method with the best proposed semi-supervised method
  (i.e., conversation model with POS tags and features).}
\end{table}

\begin{table}[b!]
\centering
\resizebox{\columnwidth}{!}{
{\tabulinesep=2mm
\begin{tabu}{ r  c  c  c  c  c  c }
\cline{2-7}
& \multicolumn{3}{c}{JeepForum} & \multicolumn{3}{c}{BenzWorld} \\
\hline
\textbf{Category} & \textbf{P} & \textbf{R} & \textbf{F} & \textbf{P} & \textbf{R} & \textbf{F} \\
\hline
\textit{Problem} & 0.58 (0.81) & 0.70 (0.61) & 0.63 (0.69) & 0.59 (0.91) & 0.72 (0.63) & 0.65 (0.74) \\

\textit{Solution} & 0.55 (0.66) & 0.72 (0.23) & 0.63 (0.34) & 0.58 (0.45) & 0.67 (0.88) & 0.62 (0.60) \\

\textit{Clarification-Req} & 0.20 (0.00) & 0.08 (0.00) & 0.12 (0.00) & 0.14 (0.00) & 0.07 (0.00) & 0.10 (0.00) \\

\textit{Clarification} & 0.13 (0.00) & 0.04 (0.00) & 0.06 (0.00) & 0.04 (0.00) & 0.01 (0.00) & 0.02 (0.00) \\

\textit{Feedback} & 0.27 (0.00) & 0.24 (0.00) & 0.26 (0.00) & 0.33 (0.00) & 0.26 (0.00) & 0.29 (0.00) \\

\textit{Other} & 0.63 (0.37) & 0.47 (0.86) & 0.54 (0.52) & 0.50 (0.30) & 0.41 (0.15) & 0.45 (0.20) \\

\textit{Micro-average} & 0.54 (0.45) & 0.54 (0.45) & 0.54 (0.45) & 0.53 (0.50) & 0.53 (0.50) & 0.53 (0.50) \\

\textit{Macro-average} & 0.38 (0.31) & 0.38 (0.28) & 0.38 (0.29) & 0.36 (0.28) & 0.36 (0.28) & 0.36 (0.28) \\
\hline
\end{tabu}
}
}
\caption{\label{table:semi-supervised-fine-grained-results} Experimental
  results of semi-supervised conversation model with POS tags and
  features for one of the folds in a 5-fold cross-validation setup
  (with the corresponding results of the best performing problem-solution heuristic in parentheses).}
\end{table}

\begin{table}[t!]
\centering
{\tabulinesep=2mm
\begin{tabu}{ r r c  c  c  c  c  c }
& & \multicolumn{6}{c}{Predicted} \\
& & \textbf{P} & \textbf{S} & \textbf{C-R} & \textbf{C} &
\textbf{F} & \textbf{O} \\
\cline{3-8}
\multirow{6}{*}{\rotatebox[origin=c]{90}{Actual}}
& \multicolumn{1}{r|}{\textbf{P}} & 342 & 70 & 5 & 2 & 26 & 45 \\
& \multicolumn{1}{r|}{\textbf{S}} & 66 & 782 & 21 & 2 & 37 & 161 \\
& \multicolumn{1}{r|}{\textbf{C-R}} & 25 & 86 & 14 & 0 & 14 & 22 \\
& \multicolumn{1}{r|}{\textbf{C}} & 11 & 42 & 4 & 1 & 23 & 16 \\
& \multicolumn{1}{r|}{\textbf{F}} & 57 & 127 & 4 & 3 & 78 & 68 \\
& \multicolumn{1}{r|}{\textbf{O}} & 93 & 364 & 16 & 13 & 94 & 485 \\
\end{tabu}
}
\caption{\label{table:confusion-matrix} Confusion matrix of the
  semi-supervised conversation model with POS tags and features, for
  one of the folds in a 5-fold cross-validation setup using the JeepForum dataset (P: Problem; S: Solution; C-R: Clarification-Request; C: Clarification; F: Feedback; O: Other).}
\end{table}

\section{Category-wise Performance and Error Analysis}
Table \ref{table:semi-supervised-fine-grained-results} shows the category-wise performance
of one of the runs of 5-fold cross-validation for both the JeepForum
and Benzworld datasets using the semi-supervised conversation model with POS tags and features. This
method outperformed the problem-solution heuristic baseline for every category except
\textit{Problem}. Table \ref{table:confusion-matrix} shows the
confusion matrix of the same experimental fold using the JeepForum dataset. The confusion matrix
 for the BenzWorld dataset is similar. The most common error was the
prediction of a non-\textit{Solution} category as \textit{Solution},
indicating a bias of the method towards predicting the majority
category. This also happened in the case of \textit{Other}, but to a
lesser extent. In addition, \textit{Clarification-Request} was often
predicted as \textit{Problem}, and \textit{Clarification} was predicted as
\textit{Solution}. This seems to occur because \textit{Clarification-Request} and
\textit{Clarification} can be understood as specific types of
\textit{Problem} and \textit{Solution} posts. Overall, the predictions of minority categories are not
practically useful, because they were less accurate than the
predictions using the random baseline. Since previous literature ignores the
analysis of category-wise performance altogether, a direct comparison
is not possible.

In order to analyze the performance of only the \textit{Problem} and
\textit{Solution} categories, another setup was used where all other categories
were coalesced into \textit{Other}. Results of the coarse-grained classification setup are
shown in Table \ref{table:semi-supervised-coarse-grained-results}. As compared to the
fine-grained classification setup, there is no significant change in $F_1$-measure
values for \textit{Problem} and \textit{Solution}. Also, this setup
performed only as well as or slightly worse than the corresponding best problem-solution
heuristic. However, the performance was much better than the baseline for \textit{Solution}.

\begin{table}[h!]
\centering
\resizebox{\columnwidth}{!}{
{\tabulinesep=2mm
\begin{tabu}{r c c c c c c }
\cline{2-7}
& \multicolumn{3}{c}{JeepForum} & \multicolumn{3}{c}{BenzWorld} \\
\hline
\textbf{Category} & \textbf{P} & \textbf{R} & \textbf{F} & \textbf{P} & \textbf{R} & \textbf{F} \\
\hline
\textit{Problem} & 0.56 (0.81) & 0.72 (0.61) & 0.63 (0.69) & 0.59 (0.91) & 0.69 (0.63) & 0.64 (0.74) \\

\textit{Solution} & 0.56 (0.66) & 0.74 (0.23) & 0.63 (0.34) & 0.59 (0.75) & 0.67 (0.27) & 0.63 (0.40) \\

\textit{Other} & 0.75 (0.60) & 0.54 (0.89) & 0.63 (0.71) & 0.63 (0.52) & 0.50 (0.91) & 0.56 (0.66) \\

\textit{Micro-average} & 0.63 (0.63) & 0.63 (0.63) & 0.63 (0.63) & 0.60 (0.61) & 0.60 (0.61) & 0.60 (0.61) \\

\textit{Macro-average} & 0.62 (0.69) & 0.66 (0.57) & 0.64 (0.63) & 0.61 (0.73) & 0.62 (0.60) & 0.61 (0.66) \\
\hline
\end{tabu}
}
}
\caption{\label{table:semi-supervised-coarse-grained-results}
  Experimental results of the semi-supervised conversation model with POS
  tags and features for one of the folds in a 5-fold cross-validation
  setup by considering all categories except \textit{Problem} and
  \textit{Solution} as the \textit{Other} category (with the corresponding results of the best performing problem-solution heuristic in parentheses).}
\end{table}

\begin{figure}[h!]
\centering
\includegraphics[width=110mm]{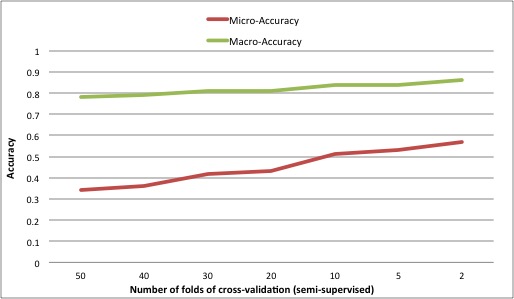}
\caption{ \label{fig:semi-supervised-no-of-folds} Performance of the
  semi-supervised conversation model with POS tags and
  features vs. the number of folds in a cross-validation setup using the
  JeepForum dataset.}
\end{figure}

\section{Effect of the Amount of Training Data}
One important measure of the quality of a learning algorithm is
whether its performance increases with increasing amount of
training data. To evaluate this, multiple \textit{n}-fold
cross-validation experiments were conducted with decreasing value of
\textit{n}, i.e., increasing number of training threads. Figure
\ref{fig:semi-supervised-no-of-folds} demonstrates that the
performance of the semi-supervised conversation
model with POS tags and features on the JeepForum dataset increased as the number
of folds decreased. The micro-accuracy value is 0.50 using 10-fold cross-validation,
which demonstrates that around 9 labeled
threads\footnote{The JeepForum dataset contains 93 threads, only
$1/10^{\text{th}}$ of which were used in a single fold of 10-fold
cross-validation.} are enough to reasonably predict categories for
unseen threads. Similar effects are seen in the case of other datasets; but to
avoid redundancy, they are not shown here. The same method in a fully
supervised setup performed even better. Table
\ref{table:comparison-supervised} shows its performance in a 10-fold
cross-validation setup as compared to an
equivalent setup that used SVMs. Despite the fact that the latter used many more features as
well as feature selection\footnote{Feature selection was done using
  information gain based attribute evaluation.}, its performance was
similar. The features used in both cases were previously described in section 3.

\begin{table}[h!]
\centering
{\tabulinesep=2mm
\begin{tabu}{r c c c }
\hline
\textbf{Category} & \textbf{P} & \textbf{R} & \textbf{F} \\
\hline
\textit{Problem} & 0.69 (0.68) & 0.71 (0.72) & 0.70 (0.70) \\

\textit{Solution} & 0.61 (0.59) & 0.77 (0.74) & 0.68 (0.66) \\

\textit{Clarification-Request} & 0.30 (0.44) & 0.21 (0.48) & 0.24 (0.46) \\

\textit{Clarification} & 0.20 (0.00) & 0.08 (0.00) & 0.12 (0.00) \\

\textit{Feedback} & 0.36 (0.59) & 0.42 (0.32) & 0.48 (0.42) \\

\textit{Other} & 0.71 (0.61) & 0.55 (0.58) & 0.62 (0.59) \\

\textit{Macro-average} & 0.48 (0.49) & 0.46 (0.47) & 0.47 (0.47) \\
\hline
\end{tabu}
}
\caption{\label{table:comparison-supervised} Experimental results of the fully
  supervised conversation model with POS tags and features in a
  10-fold cross-validation setup using the JeepForum dataset (with corresponding results using SVMs in parentheses).}
\end{table}

\section{Summary of Experimental Results}
Experimental results indicate that purely unsupervised
methods are not adequate for tackling a task as complex as forum post
categorization. However, they are able to capture some sequential
dependencies, as observed from the fact that they outperformed two
trivial baselines (i.e., the random and majority baselines). Using post
embeddings (which is still purely unsupervised), the performance did
not conclusively improve. But knowledge of POS tags and simple textual
features provided more context for classification, and thus, enabled the
technique to classify more accurately. 

The novel proposal of incorporating a few labeled examples for initializing the
model priors led to better performance than the problem-solution
heuristic baselines in most cases. Direct comparison with existing
methods is not possible due to limitations in availability of data
and code. However, approximate comparison setups demonstrate the
better performance of proposed methods. Prediction of \textit{Problem}
and \textit{Solution} categories were the most accurate, followed by
\textit{Other} and \textit{Feedback}. However, predictions of the minority
categories, \textit{Clarification-Request} and
\textit{Clarification}, were not accurate enough to be practically
useful, since the maximum accuracy value is 0.30.

\chapter{Conclusions and Future Work}
\section{Conclusions}
This paper described the problem of forum post categorization, and discussed the need for automatic methods to solve it. The relevant previous work was presented and an argument was made for the need for unsupervised and semi-supervised methods to solve the problem. Subsequently, methods were proposed for categorizing forum posts using sequence models, which distinguish between categories, using language models based on word and part-of-speech probability distributions, in addition to manually specified features. The unsupervised methods include the novel application of conversation models that were previously proposed for other tasks. Although the experimental results demonstrate that they are not practically useful, they are shown to perform better than previously proposed methods. Hence, it can be safely concluded that the current unsupervised methods are not robust enough to capture the complexity of forum post categorization. Next, it was proposed to use a novel semi-supervised version of the earlier methods by employing a few labeled threads to guide the process. Experimental results demonstrate that these methods outperformed all the baselines. Also, an indirect comparison with a semi-supervised method proposed in previous work, demonstrates better performance.

\section{Future Work}
Discussion forum posts often contain multiple dialogue categories , i.e., a post could start with a \textit{Solution} to a previous \textit{Problem}, and end with a new \textit{Problem} posed for users to discuss in future posts. In such cases, the post is annotated with a single representative category. Although this might be straightforward for human annotators, the proposed methods have no intuition about this. Hence, it might be useful to employ summarization, so as to retain the overall meaning of the post, and cut out the parts that are not representative. Such methods need only classify the relevant text in the post and might perform better. This problem could also be tackled by classifying individual sentences in posts, rather than the post as a whole. This could be done in a two-tier HMM setup where the first level comprises sentence classification, and the second level comprises of post classification. However, this proposal is dependent on the availability of datasets that are annotated by category at the sentence level. Instead, majority voting or other heuristics could be employed to pool the predicted categories of individual sentences into a single post category.

Since all the proposed methods employ first-order HMMs, they lack the knowledge of long-range dependencies between different categories. Consequently, they are unable to learn that a post can not be classified as \textit{Solution}, without a \textit{Problem} post before it. This problem can be addressed by using higher-order Markov chains, but it would lead to much greater run-time and space complexity. Instead, the use of heuristics to flag certain categories, based on prior post categories in the thread, could resolve this problem more efficiently.

Comparison of fine-grained and coarse-grained classification results indicates that \textit{Clarification-Request} and \textit{Clarification} categories are not easy to identify. Since inter-annotator agreement values for these two categories are also the least among all categories, it seems that manual identification is also not easy. Hence, future work should either discard these categories or use them for annotation in a controlled setting with trained expert annotators, as opposed to crowdsourcing.

\clearpage

\addcontentsline{toc}{chapter}{Bibliography}
\bibliographystyle{plainnat}
\bibliography{ut-thesis}


\end{document}